\documentclass{article}

\usepackage{arxiv}

\usepackage[utf8]{inputenc} 
\usepackage[T1]{fontenc}    
\usepackage{hyperref}       
\usepackage{url}            
\usepackage{booktabs}       
\usepackage{amsfonts}       
\usepackage{nicefrac}       
\usepackage{microtype}      
\usepackage{lipsum}		
\usepackage{graphicx}
\usepackage{natbib}
\usepackage{doi}
\usepackage{amsmath}
\usepackage{tikz}
\usepackage{geometry}
\usepackage{tikz}
\usetikzlibrary{arrows.meta, positioning, shapes.geometric}
\usepackage{amsmath, amssymb}

\usepackage[most]{tcolorbox}

\usepackage{listings}

\usepackage{microtype}

\usepackage{hyperref}

\usetikzlibrary{mindmap, shadows, positioning, backgrounds}
\usetikzlibrary{positioning, shapes, arrows.meta}
\usepackage{amsmath, amssymb}

\title{YOLOE-26: Integrating YOLO26 with YOLOE for Real-Time Open-Vocabulary Instance Segmentation}

\author{
  \href{https://orcid.org/0000-0002-5417-6744}{\includegraphics[scale=0.06]{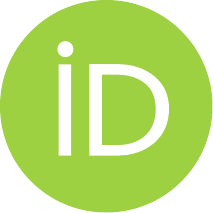}\hspace{1mm}Ranjan Sapkota\textsuperscript{}} \quad
  \href{https://orcid.org/0000-0001-5337-4848}{\includegraphics[scale=0.06]{orcid.pdf}\hspace{1mm}Manoj Karkee\textsuperscript{}} \\
  \textsuperscript{}Cornell University, Biological \& Environmental Engineering, Ithaca, NY 14850, USA \\
  \texttt{rs2672@cornell.edu} \\
}

\renewcommand{\shorttitle}{YOLOE-26:  (\href{https://docs.ultralytics.com/models/yolo26/ }{Ultralytics YOLO26 Official Source Link})}

\hypersetup{
pdftitle={A template for the arxiv style},
pdfsubject={q-bio.NC, q-bio.QM},
pdfauthor={David S.~Hippocampus, Elias D.~Striatum},
pdfkeywords={First keyword, Second keyword, More},
}
\begin{document}
\maketitle

\begin{abstract}
This paper presents YOLOE-26, a unified framework that integrates the deployment-optimized YOLO26 architecture with the open-vocabulary learning paradigm of YOLOE for real-time open-vocabulary instance segmentation. Building on the NMS-free, end-to-end design of YOLOv26, the proposed approach preserves the hallmark efficiency and determinism of the YOLO family while extending its capabilities beyond closed-set recognition. YOLOE-26 employs a convolutional backbone with PAN/FPN-style multi-scale feature aggregation, followed by end-to-end regression and instance segmentation heads. A key architectural contribution is the replacement of fixed class logits with an object embedding head, which formulates classification as similarity matching against prompt embeddings derived from text descriptions, visual examples, or a built-in vocabulary. To enable efficient open-vocabulary reasoning, the framework incorporates Re-Parameterizable Region-Text Alignment (RepRTA) for zero-overhead text prompting, a Semantic-Activated Visual Prompt Encoder (SAVPE) for example-guided segmentation, and Lazy Region Prompt Contrast for prompt-free inference. All prompting modalities operate within a unified object embedding space, allowing seamless switching between text-prompted, visual-prompted, and fully autonomous segmentation. Extensive experiments demonstrate consistent scaling behavior and favorable accuracy--efficiency trade-offs across model sizes in both prompted and prompt-free settings. The training strategy leverages large-scale detection and grounding datasets with multi-task optimization and remains fully compatible with the Ultralytics ecosystem for training, validation, and deployment. Overall, YOLOE-26 provides a practical and scalable solution for real-time open-vocabulary instance segmentation in dynamic, real-world environments.
\end{abstract}

\begin{figure}[h!]
     \centering
     \includegraphics[width=0.80 \linewidth]{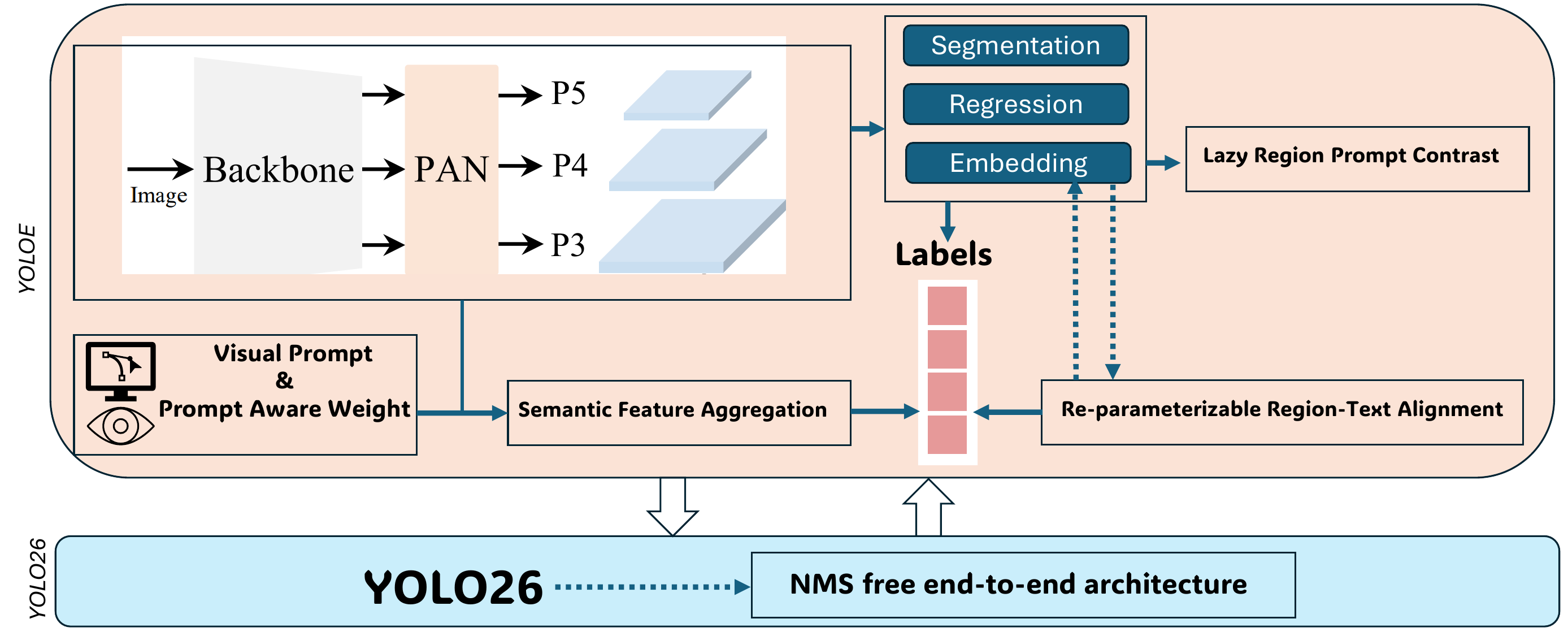}
    \caption{Bird’s-eye diagram: simplified architecture of YOLOE-26 for open-vocabulary instance segmentation.}
    \label{fig:one}
\end{figure}

\section{INTRODUCTION}

Object detection and instance segmentation constitute two of the most fundamental problems in computer vision, enabling machines to localize, recognize, and delineate objects within images and video streams. These capabilities form the core of numerous real-world applications, including autonomous driving, robotics, intelligent surveillance, medical image analysis, precision agriculture, and smart manufacturing \cite{zhao2019object, zou2023object, rana2024artificial, khan2025objectdetection}. In such applications, real-time inference, low latency, and deployment efficiency are often as critical as recognition accuracy.

The You Only Look Once (YOLO) family has fundamentally shaped modern real-time object detection by introducing unified, single-stage detection pipelines. YOLOv1 reframed detection as an end-to-end regression task, enabling unprecedented inference speed \cite{redmon2016you}. YOLOv2 and YOLOv3 extended this paradigm through anchor box clustering, multi-scale training, deeper Darknet backbones, and residual feature fusion, significantly improving robustness and small-object detection \cite{redmon2017yolo9000, redmon2018yolov3}. Subsequent versions emphasized efficiency and stability, with YOLOv4 adopting CSPDarknet and Mish activation \cite{bochkovskiy2020yolov4}, YOLOv5 transitioning to PyTorch with modern training pipelines \cite{sapkota2025ultralytics}, YOLOv6 introducing EfficientRep and anchor-free heads \cite{li2022yolov6}, and YOLOv7 leveraging re-parameterized ELAN architectures \cite{wang2023yolov7}. Recent models reflect a shift toward end-to-end and attention-aware designs, including anchor-free YOLOv8 \cite{sohan2024review, wang2023uav}, PGI-enhanced YOLOv9 \cite{wang2024yolov9}, NMS-free YOLOv10 \cite{wang2024yolov10}, extended-task YOLO11, and attention- and graph-based YOLOv12 and YOLOv13 \cite{tian2025yolov12, lei2025yolov13}.

The release of YOLOv26 in 2025 represents a culmination of these trends. Rather than increasing architectural complexity, YOLOv26 adopts a deployment-first philosophy centered on efficiency, robustness, and simplicity. Key innovations include a native NMS-free end-to-end predictor, removal of Distribution Focal Loss (DFL) for faster inference, and the introduction of the MuSGD optimizer for stable and rapid convergence \cite{sapkota2025yolo26}. These design choices significantly reduce end-to-end latency and improve performance on low-power CPUs and edge devices, while supporting multiple vision tasks such as detection, instance segmentation, pose estimation, oriented detection, and classification.

Despite the substantial gains in accuracy-latency efficiency demonstrated by successive YOLO generations in Fig.~\ref{fig:yolograph}(a,b), including the end-to-end, NMS-free design of YOLOv26, these models remain inherently constrained by a \emph{closed-vocabulary} formulation in which object categories are fixed at training time and cannot adapt to unseen concepts at inference. This limitation poses a significant challenge in open-world scenarios, where object categories continuously evolve and retraining is impractical. Recent advances in foundation models and vision-language learning have given rise to \emph{open-vocabulary object detection and instance segmentation}, enabling models to recognize unseen categories through text prompts, visual examples, or prompt-free inference \cite{shi2025harnessing, sapkota2026object}. However, many existing open-vocabulary approaches rely on transformer-heavy architectures or large language models, resulting in high computational cost, slow inference, and limited deployability on edge hardware \cite{sapkota2026object}.

The YOLOE (You Only Look Once - Everything) paradigm addresses this gap by extending the YOLO framework with embedding-based classification and unified support for text prompts, visual prompts, and prompt-free operation, enabling ``seeing anything'' while preserving YOLO’s hallmark efficiency \cite{wang2025yoloe}. By aligning visual features with semantic embeddings instead of fixed class logits, YOLOE introduces foundation-model-inspired open-vocabulary learning into real-time detection and segmentation.

In this paper, we present a comprehensive evaluation of YOLOE-26,which combines the NMS-free, end-to-end detection pipeline of YOLOv26 with the open-vocabulary learning mechanisms of YOLOE, enabling real-time instance segmentation across diverse prompting paradigms(\href{https://docs.ultralytics.com/models/yolo26/}{Source Link}). By combining the deployment-optimized design of YOLOv26 with the open-vocabulary capabilities of YOLOE, YOLOE-26 establishes a unified and practical framework for real-time, open-world instance segmentation. This study systematically analyzes its performance across text-prompted, visual-prompted, and prompt-free settings, highlighting its accuracy–efficiency trade-offs and suitability for next-generation edge and open-world vision systems.

\subsection{Background and Motivation}
Convolutional neural network (CNN)–based object detection frameworks, particularly the YOLO family, have dominated real-time visual perception for nearly a decade due to their unified architectures, high inference speed, and strong accuracy–efficiency trade-offs \cite{ramos2025decade, bai2025decade}. From YOLOv1 to YOLOv26, these models progressively transformed object detection from grid-based regression into highly optimized end-to-end pipelines. Early variants (YOLOv1–YOLOv3) relied on dense grid predictions with fixed-category classification heads, which limited semantic flexibility and required careful anchor and scale tuning \cite{redmon2016you, redmon2017yolo9000, redmon2018yolov3}. Later generations incorporated multi-scale feature pyramids, deeper backbones, and improved loss formulations to enhance robustness across object sizes, establishing YOLO as the de facto standard for real-time detection in resource-constrained environments \cite{bochkovskiy2020yolov4}.

\begin{figure}[h!]
     \centering
     \includegraphics[width=0.70\linewidth]{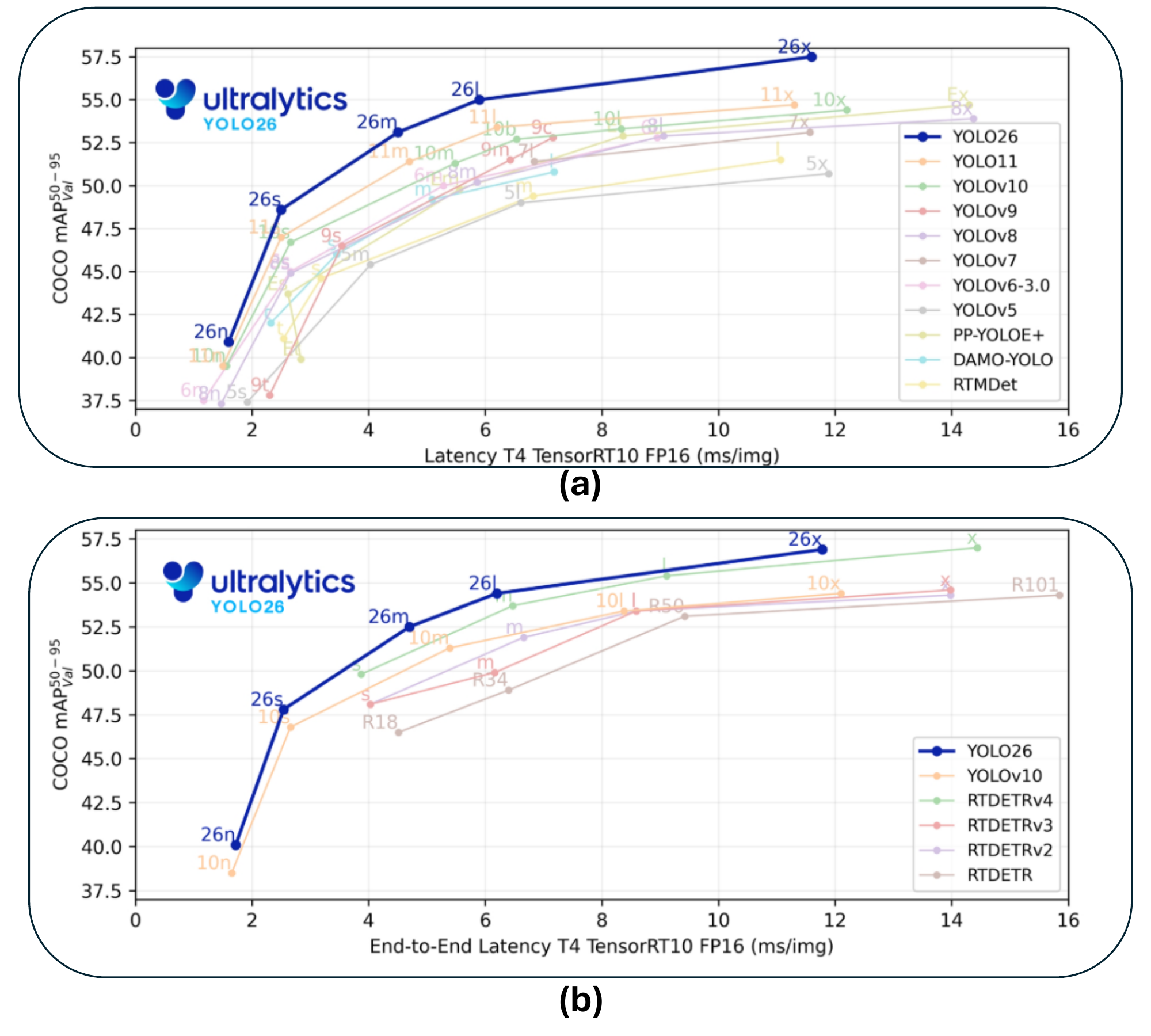}
    \caption{Performance comparison of YOLO26 under TensorRT FP16 on an NVIDIA T4 GPU  (\href{https://docs.ultralytics.com/models/yolo26/}{Source Link}). (a) COCO mAP(50–95) versus inference latency (ms/image), comparing YOLO26 with earlier YOLO versions and other real-time detectors, highlighting its improved accuracy–speed trade-off. (b) COCO mAP(50–95) versus end-to-end latency, comparing YOLO26 with YOLOv10 and RT-DETR variants, illustrating its advantage in overall pipeline efficiency.}
    \label{fig:yolograph}
\end{figure}

As YOLO matured, architectural focus shifted toward deployment robustness and pipeline simplification. YOLOv5–YOLOv7 introduced PyTorch-based implementations, anchor-free detection heads, re-parameterizable convolutional blocks, and efficient feature aggregation mechanisms, significantly reducing training and inference complexity \cite{sapkota2025ultralytics, li2022yolov6, wang2023yolov7}. More recent versions, including YOLOv8 and YOLOv9, emphasized decoupled heads, task-aligned optimization, and multi-task perception, extending YOLO’s applicability to instance segmentation, pose estimation, and panoptic understanding \cite{sohan2024review, wang2024yolov9}. This evolution culminated in YOLOv10 and YOLOv26, which eliminated heuristic post-processing such as non-maximum suppression (NMS), enabling fully end-to-end detection with reduced latency and improved determinism \cite{wang2024yolov10, sapkota2025yolo26}. As illustrated in Fig.~\ref{fig:yolograph}(a), YOLOv26 achieves a superior accuracy–latency balance compared to earlier YOLO variants and other real-time detectors, while Fig.~\ref{fig:yolograph}(b) highlights its advantage in end-to-end pipeline efficiency relative to transformer-based real-time baselines.

Despite these architectural advances, CNN-based YOLO detectors, including YOLOv26 remain fundamentally constrained by a closed-set learning paradigm, where object categories are predefined during training and fixed at inference time \cite{loukovitis2025model, wang2025slbdetection}. In real-world scenarios such as autonomous robotics, agricultural monitoring, and industrial inspection, object categories frequently evolve, making repeated data collection, retraining, and redeployment impractical. These limitations hinder adaptability in open-world environments and motivate the transition toward open-vocabulary vision systems.

Open-vocabulary detection and instance segmentation methods attempt to overcome closed-set constraints by leveraging large-scale vision–language pretraining and semantic embeddings \cite{weijler2025openhype, wu2023tidybot}. Models such as GLIP, Grounding DINO, OWL-ViT, DINO-X, X-Decoder, OpenSeeD, and SEEM demonstrate strong zero-shot and open-set capabilities by aligning visual regions with textual or multimodal representations. However, these approaches typically rely on transformer-heavy backbones, dense cross-modal attention, and external language models, resulting in high computational cost, increased inference latency, and large memory footprints. Such characteristics severely limit real-time performance and edge deployability, particularly for safety-critical and low-power applications.

\begin{figure}[h!]
     \centering
     \includegraphics[width=0.70\linewidth]{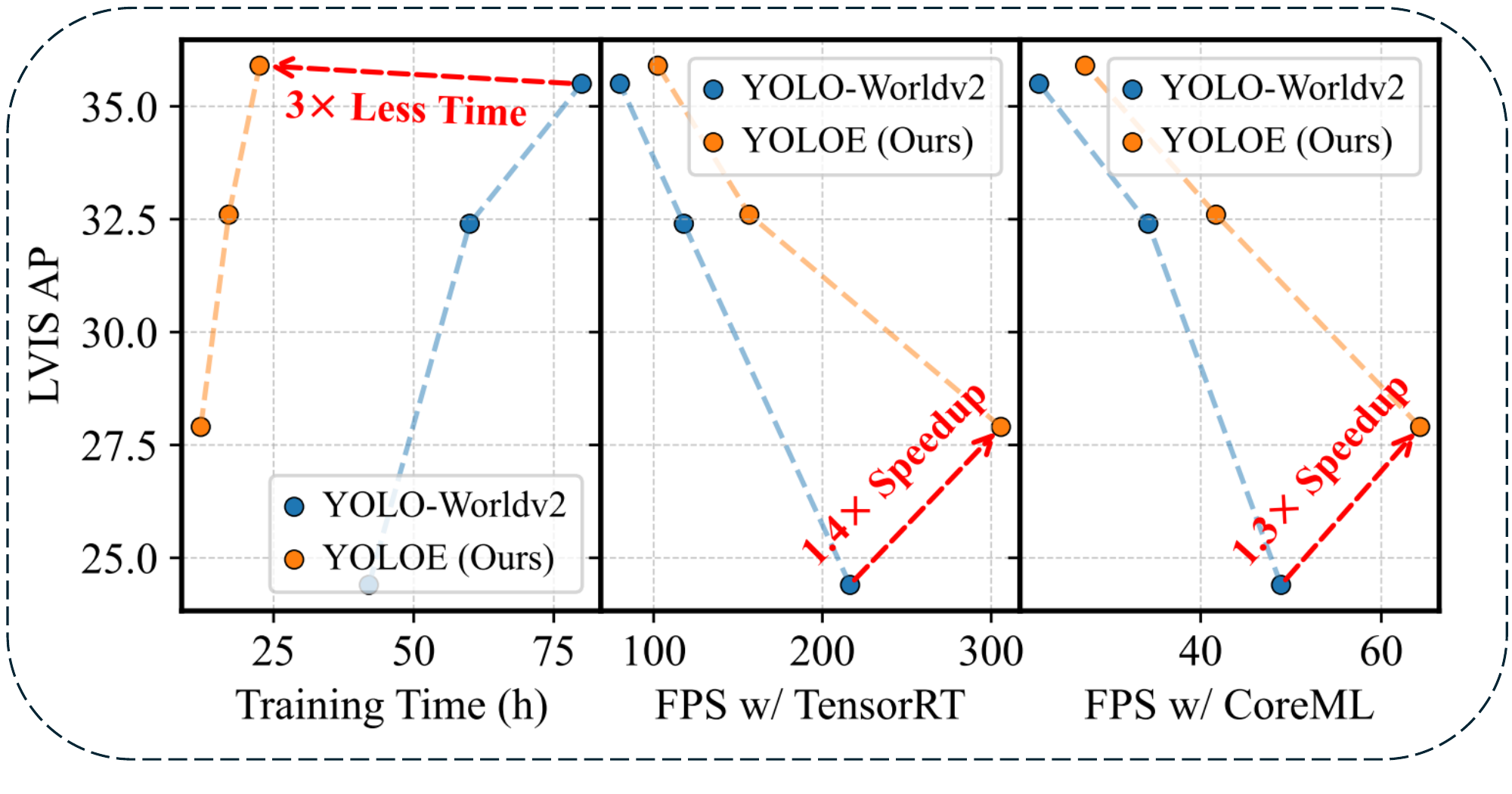}
    \caption{Comparison of performance, training cost, and inference efficiency between YOLOE and advanced YOLO-Worldv2 in terms of open text prompts. LVIS AP is evaluated on minival set and FPS w/ TensorRT and w/ CoreML is measured on T4 GPU and iPhone 12, respectively. The results highlight our superiority. (Source: YOLOE paper \cite{wang2025yoloe}}
    \label{fig:yoloegraph}
\end{figure}

YOLOE represents a critical step toward integrating open-vocabulary learning into efficient YOLO-style architectures \cite{wang2025yoloe}. By introducing embedding-based classification and unified support for text prompts, visual prompts, and prompt-free inference, YOLOE enables open-vocabulary detection and segmentation within a single model \cite{jose2026zero, liu2025yoloe, palaniappan2025yolo}. Nevertheless, early YOLOE designs still exhibit limitations related to prompt handling efficiency, scalability across deployment scenarios, and full exploitation of end-to-end detection pipelines. As reflected in Fig.~\ref{fig:yoloegraph}, while YOLOE improves open-vocabulary performance compared to prior YOLO-World variants, challenges remain in balancing training cost, inference efficiency, and real-world deployability.

The motivation behind YOLOE-26 is to systematically address these limitations by tightly integrating YOLOv26’s NMS-free, end-to-end detection framework with the open-vocabulary learning mechanisms of YOLOE. By unifying deployment-efficient CNN-based detection with lightweight vision-language embedding strategies, YOLOE-26 enables text-prompted, visual-prompted, and prompt-free instance segmentation without sacrificing real-time performance. This design positions YOLOE-26 as a practical solution for dynamic, open-world vision applications, including robotics, autonomous systems, surveillance, and precision agriculture, where both semantic flexibility and deployment efficiency are essential.

\section{YOLOE-26 Architecture Overview}
\subsection{Core YOLO26 Architectural Backbone and End-to-End Design}

\begin{figure}[h!]
     \centering
     \includegraphics[width=0.99\linewidth]{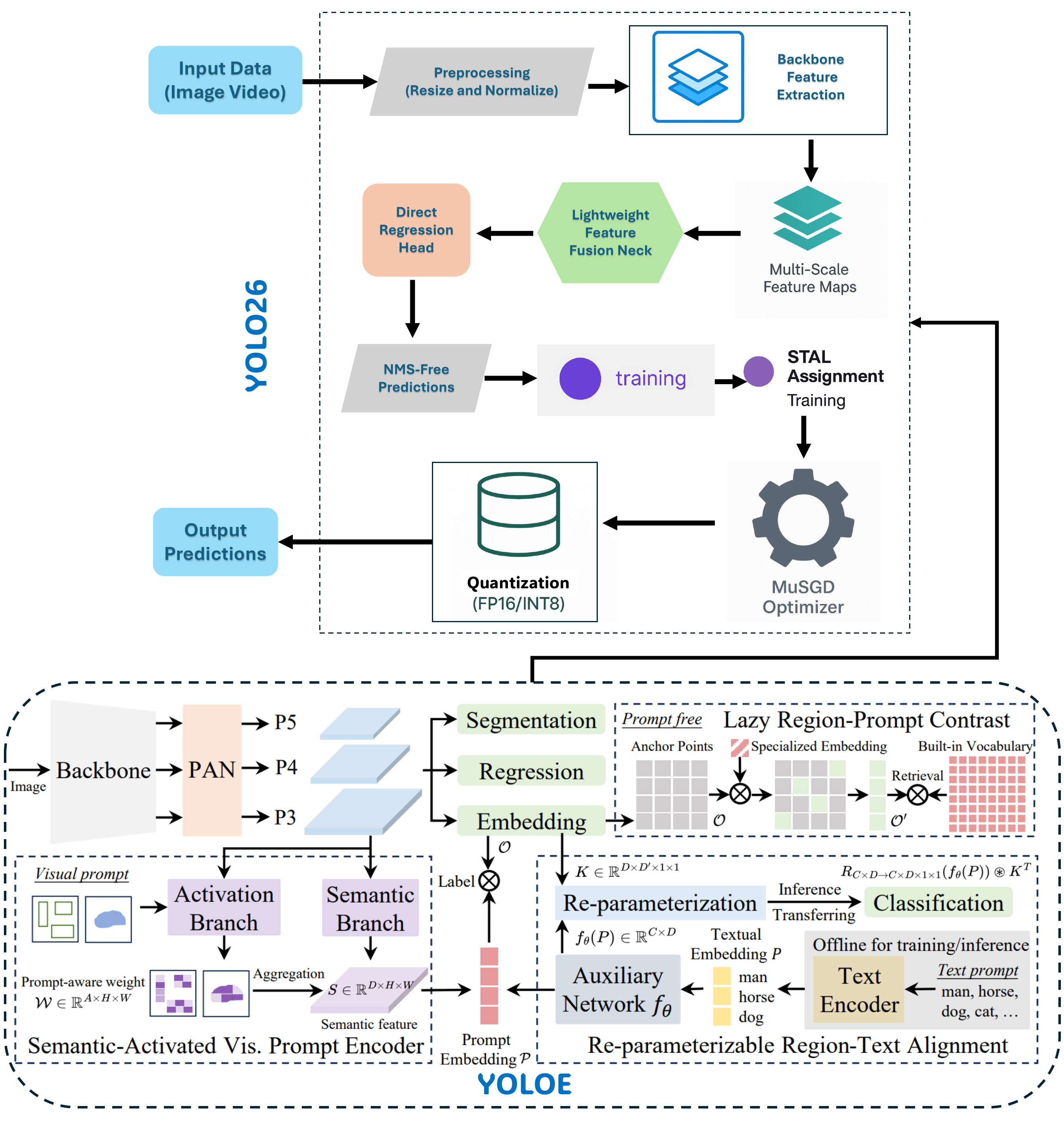}
    \caption{Architectural diagram of YOLOE-26 for open-vocabulary instance segmentation. The upper part illustrates the core YOLOv26 end-to-end detection and segmentation pipeline, while the lower part depicts the YOLOE (\cite{wang2025yoloe}) components that enable text-prompted, visual-prompted, and prompt-free open-vocabulary learning. }
    \label{fig:archite}
\end{figure}

YOLOE-26 is a unified architecture that tightly integrates the deployment-efficient, NMS-free design of YOLOv26 with the open-vocabulary learning mechanisms introduced in YOLOE. As illustrated in Fig.~\ref{fig:archite}, the model follows the canonical YOLO pipeline backbone, neck, and task-specific heads while replacing the conventional closed-set classification head with a semantic embedding formulation that supports open-world instance segmentation.

\paragraph{YOLOv26 Backbone and Feature Extraction:}
At its core, YOLOE-26 inherits the convolutional backbone of YOLOv26, which is designed for efficient multi-scale feature extraction across diverse hardware platforms. Given an input image $I \in \mathbb{R}^{3 \times H \times W}$, the backbone applies a hierarchy of convolutional layers to extract feature maps at multiple resolutions. These features encode both low-level spatial details and high-level semantic context, which are essential for detecting objects of varying sizes. Compared to earlier YOLO variants, YOLOv26 emphasizes simplified convolutional blocks and optimized gradient flow, reducing computational overhead while preserving representational capacity.

\paragraph{Neck: PAN/FPN-Style Feature Aggregation:}
The extracted backbone features are passed to a PAN/FPN-style neck, which aggregates information across scales. Let $\{P_3, P_4, P_5\}$ denote feature maps at increasing semantic levels and decreasing spatial resolutions. The neck performs top-down and bottom-up fusion through upsampling, concatenation, and convolution operations, ensuring that each detection point has access to both fine-grained localization cues and global semantic information. This multi-scale aggregation is particularly important for instance segmentation, where precise object boundaries must be inferred alongside object identity.

\paragraph{End-to-End Regression and Segmentation Heads:}
For each anchor point (or grid location) in the aggregated feature maps, YOLOE-26 employs multiple task-specific heads. The regression head predicts bounding box parameters, typically encoded as offsets relative to the anchor point, enabling precise object localization. In parallel, the instance segmentation head follows the prototype-based design common to modern YOLO segmentation models. It produces a set of global mask prototypes and per-instance mask coefficients, which are linearly combined to generate instance-specific segmentation masks. This design decouples spatial mask representation from instance prediction, achieving high efficiency and scalability.

\paragraph{NMS-Free End-to-End Detection:}
A defining characteristic of YOLOv26, inherited by YOLOE-26, is the removal of non-maximum suppression (NMS). Traditional YOLO pipelines rely on NMS as a post-processing step to eliminate redundant detections, introducing additional latency and heuristic complexity. YOLOv26 instead adopts an end-to-end training formulation that enforces consistent assignment between predictions and ground truth, allowing the network to learn mutual exclusivity directly. As a result, the final predictions are obtained in a single forward pass, improving determinism, reducing latency, and simplifying deployment an advantage that becomes increasingly important in open-vocabulary settings with large and dynamic category spaces.

\paragraph{Object Embedding Head for Open-Vocabulary Learning:}
The most critical architectural modification introduced by YOLOE-26 is the replacement of the closed-set classification head with an object embedding head. Instead of predicting logits over a fixed set of class labels, the object embedding head outputs a semantic embedding vector for each anchor point. Formally, let $\mathcal{O} \in \mathbb{R}^{N \times D}$ denote the object embeddings produced for $N$ anchor points, where $D$ is the embedding dimension. These embeddings represent visual object instances in a shared semantic space, enabling flexible matching with arbitrary category representations.

\paragraph{Prompt Embeddings and Similarity-Based Classification:}
YOLOE-26 supports three prompting modes text prompts, visual prompts, and prompt-free inference by encoding all prompts into a common embedding space. Given a set of $C$ prompts, their embeddings are denoted as $\mathcal{P} \in \mathbb{R}^{C \times D}$. Category prediction is then formulated as a similarity operation between object embeddings and prompt embeddings:
\begin{equation}
\text{Label} = \mathcal{O} \cdot \mathcal{P}^{T} \in \mathbb{R}^{N \times C},
\end{equation}
where each entry represents the affinity between an anchor point and a prompt. This formulation replaces fixed-category classification with a flexible retrieval-style mechanism, enabling zero-shot recognition of unseen categories.

\paragraph{Re-Parameterizable Region–Text Alignment (RepRTA):}
To improve visual–textual alignment without incurring inference overhead, YOLOE-26 employs RepRTA during training. Text prompts are first encoded using a pretrained text encoder, producing embeddings $\mathcal{P}$. A lightweight auxiliary network $f_{\theta}$ refines these embeddings during training to better align with visual features. The refined embeddings interact with object embeddings via convolutional operations:
\begin{equation}
\text{Label} = R(I \circledast K) \cdot (f_{\theta}(\mathcal{P}))^{T},
\end{equation}
where $I$ denotes intermediate feature maps and $K$ represents convolution kernels. After training, the auxiliary network is re-parameterized into the object embedding head, yielding new kernels $K'$:
\begin{equation}
K' = R(f_{\theta}(\mathcal{P})) \circledast K^{T}.
\end{equation}
This transformation preserves the standard YOLO inference path, ensuring zero additional runtime cost.

\paragraph{Semantic-Activated Visual Prompt Encoder (SAVPE):}
For visual prompting, YOLOE-26 introduces SAVPE, a lightweight encoder that avoids transformer-heavy designs. SAVPE consists of a semantic branch that extracts prompt-agnostic semantic features and an activation branch that generates prompt-aware weights from visual cues such as bounding boxes or masks. These components are aggregated to form visual prompt embeddings:
\begin{equation}
\mathcal{P} = \text{Concat}(G_1, \ldots, G_A), \quad
G_i = \mathcal{W}_{i:i+1} \cdot S^{T}_{\frac{D}{A}i:\frac{D}{A}(i+1)},
\end{equation}
where $S$ denotes semantic features and $\mathcal{W}$ represents activation weights. This grouped formulation achieves efficient prompt encoding with minimal computational overhead.

\paragraph{Lazy Region–Prompt Contrast for Prompt-Free Inference:}
In the absence of explicit prompts, YOLOE-26 adopts Lazy Region–Prompt Contrast (LRPC) to identify and name objects efficiently. A specialized prompt embedding $\mathcal{P}_s$ is trained to detect objectness, filtering relevant anchor points:
\begin{equation}
\mathcal{O}' = \{o \in \mathcal{O} \mid o \cdot \mathcal{P}_s^{T} > \delta\}.
\end{equation}
Only the filtered set $\mathcal{O}'$ is matched against a large built-in vocabulary, significantly reducing computational cost while maintaining accuracy.

\subsection{Unified Object Embedding Space}
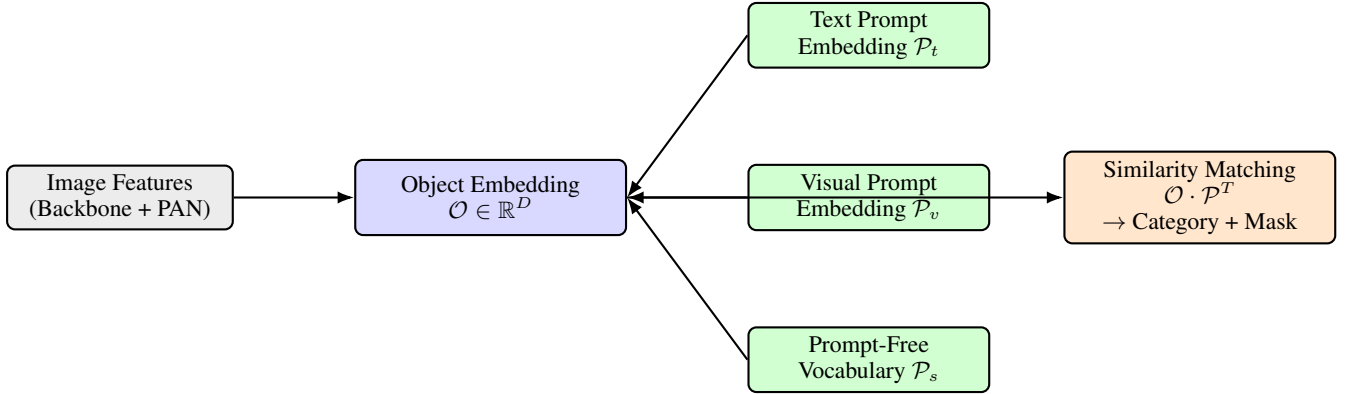
\begin{figure}[t]
\centering
\begin{tikzpicture}[
    font=\small,
    node distance=12mm and 16mm,
    block/.style={rectangle, rounded corners=3pt, draw, thick, align=center,
                  minimum width=30mm, minimum height=8mm, fill=gray!15},
    emb/.style={rectangle, rounded corners=3pt, draw, thick, align=center,
                minimum width=36mm, minimum height=10mm, fill=blue!15},
    promptbox/.style={rectangle, rounded corners=3pt, draw, thick, align=center,
                      minimum width=32mm, minimum height=8mm, fill=green!18},
    outputbox/.style={rectangle, rounded corners=3pt, draw, thick, align=center,
                      minimum width=36mm, minimum height=10mm, fill=orange!20},
    arrow/.style={-{Latex[length=2.2mm]}, thick}
]

\node[block] (img) {Image Features\\(Backbone + PAN)};
\node[emb, right=of img] (obj) {Object Embedding\\$\mathcal{O}\in\mathbb{R}^{D}$};

\node[promptbox, above right=of obj] (text) {Text Prompt\\Embedding $\mathcal{P}_t$};
\node[promptbox, right=of obj] (visual) {Visual Prompt\\Embedding $\mathcal{P}_v$};
\node[promptbox, below right=of obj] (pf) {Prompt-Free\\Vocabulary $\mathcal{P}_s$};

\node[outputbox, right=58mm of obj] (sim) {Similarity Matching\\$\mathcal{O}\cdot\mathcal{P}^{T}$\\$\rightarrow$ Category + Mask};

\draw[arrow] (img) -- (obj);
\draw[arrow] (obj) -- (sim);

\draw[arrow] (text.west) -- (obj.east);
\draw[arrow] (visual.west) -- (obj.east);
\draw[arrow] (pf.west) -- (obj.east);

\end{tikzpicture}
\caption{Unified object embedding space in YOLOE-26. Visual features are encoded into object embeddings and matched with text prompts, visual prompts, or prompt-free vocabulary embeddings via similarity-based inference for open-vocabulary instance segmentation.}
\label{fig:unified_embedding}
\end{figure}

As illustrated in Figure \ref{fig:unified_embedding}, a central design principle of YOLOE-26 is the replacement of conventional closed-set classification with a unified object embedding space that supports flexible, open-vocabulary reasoning. In traditional YOLO detectors, each anchor point predicts a probability distribution over a fixed set of class labels using a softmax or sigmoid-based classifier \cite{zhou2026projection, qi2023balanced, wang2025scb}. Such formulations tightly couple visual features to predefined categories, limiting generalization to unseen concepts. YOLOE-26 instead represents each detected instance through a continuous semantic embedding, enabling category inference via similarity matching rather than explicit class prediction.

Concretely, for each anchor point, the object embedding head outputs a $D$-dimensional vector that encodes the visual appearance and semantic attributes of the underlying object \cite{saini2022disentangling, zhang2022dimension}. These object embeddings are learned jointly with the detection and segmentation tasks, ensuring alignment between spatial localization, mask prediction, and semantic representation. In parallel, category descriptions provided as text prompts, visual prompts, or prompt-free object descriptors are mapped into the same $D$-dimensional embedding space, producing a set of prompt embeddings. Classification is then performed by computing the similarity typically via inner product or cosine similarity between object embeddings and prompt embeddings, yielding affinity scores that indicate the likelihood of each object belonging to a given semantic category.

This embedding-based formulation unifies multiple inference modes within a single architecture. Text prompts allow users to specify object categories using natural language descriptions, visual prompts enable category specification through example regions or masks, and the prompt-free mode relies on learned objectness embeddings to retrieve category names from a built-in vocabulary. Importantly, all three modes share the same object embedding head, eliminating the need for task-specific classification branches or external language models during inference.

From an application perspective, the unified object embedding space enables zero-shot and open-world instance segmentation without retraining. New categories can be introduced at inference time by simply providing corresponding prompt embeddings, making YOLOE-26 well suited for dynamic real-world environments such as robotics, autonomous navigation, surveillance, and precision agriculture. By decoupling semantic reasoning from fixed label spaces while preserving the efficiency of YOLO-style detection, YOLOE-26 achieves a practical balance between flexibility and real-time deployability.

\subsection{Performance Evaluation of Text-/Visual-Prompted and Prompt-Free Open-Vocabulary Instance Segmentation}

This subsection presents a comprehensive performance analysis of YOLOE-26 under two complementary open-vocabulary inference paradigms: \emph{text/visual-prompted segmentation} and \emph{prompt-free segmentation}. Quantitative results are summarized in Table~\ref{tab:yoloe26_seg_tv} and Table~\ref{tab:yoloe26_seg_pf}, respectively, both evaluated at a resolution of 640 px using end-to-end (e2e) metrics on the minival benchmark. Together, these tables provide critical insights into the accuracy–efficiency trade-offs of YOLOE-26 across model scales and prompting strategies, directly informing real-world deployment decisions.

Table~\ref{tab:yoloe26_seg_tv} reports performance when explicit semantic guidance is provided through text or visual prompts. A clear and consistent scaling trend is observed across the YOLOE-26 model family, where increasing model capacity yields substantial gains in segmentation accuracy. In particular, YOLOE-26x-seg achieves the highest overall performance, reaching a mAP$_{50\text{--}95}$ of 39.5 (text) and 36.2 (visual), alongside strong results across rare, common, and frequent category splits. This highlights the effectiveness of the unified object embedding space and prompt-aware alignment mechanisms in handling long-tailed category distributions, which are common in open-world scenarios. From an application perspective, such performance is especially valuable in domains like robotics, surveillance, and precision agriculture, where rare object instances (e.g., unusual tools, uncommon vehicle types, or early-stage crop anomalies) must be segmented reliably based on semantic descriptions rather than fixed labels.

Mid-scale models such as YOLOE-26m-seg and YOLOE-26l-seg offer a compelling balance between accuracy and computational cost. For instance, YOLOE-26l-seg delivers mAP$_{50\text{--}95}$ values exceeding 36 with fewer than 90 FLOPs (B), making it well suited for edge GPUs and embedded accelerators used in autonomous drones, mobile robots, and industrial inspection systems. In contrast, the nano and small variants prioritize efficiency, achieving respectable segmentation accuracy with significantly lower parameter counts and FLOPs, which is critical for deployment on low-power devices and real-time video analytics pipelines.

\begin{table}[h!]
\centering
\scriptsize
\setlength{\tabcolsep}{3.8pt}
\renewcommand{\arraystretch}{1.15}
\caption{YOLOE-26 open-vocabulary instance segmentation performance using \textbf{text and visual prompts} at 640 px resolution. Results are reported on the minival set with end-to-end (e2e) evaluation. Rare (r), common (c), and frequent (f) category splits follow standard open-vocabulary benchmarks. Models are trained on Objects365v1, GQA, and Flickr30k datasets.}
\label{tab:yoloe26_seg_tv}
\vspace{2pt}

\begin{tabular}{l c c c c c c c}
\toprule
\textbf{Model} & \textbf{Size} & \textbf{Prompt} & \textbf{mAP$_{50\text{--}95}$} & \textbf{mAP$_{50\text{--}95}$} & \textbf{mAP$_r$ / $c$ / $f$} & \textbf{Params} & \textbf{FLOPs} \\
 & \textbf{(px)} & \textbf{Type} & \textbf{(Text / Visual)} & \textbf{(Text / Visual)} & \textbf{(Text / Visual)} & \textbf{(M)} & \textbf{(B)} \\
\midrule
YOLOE-26n-seg & 640 & Text / Visual & 23.7 / 20.9 & 24.7 / 21.9 & 20.5 / 17.6 \quad 24.1 / 22.3 \quad 26.1 / 22.4 & 4.8 & 6.0 \\
YOLOE-26s-seg & 640 & Text / Visual & 29.9 / 27.1 & 30.8 / 28.6 & 23.9 / 25.1 \quad 29.6 / 27.8 \quad 33.0 / 29.9 & 13.1 & 21.7 \\
YOLOE-26m-seg & 640 & Text / Visual & 35.4 / 31.3 & 35.4 / 33.9 & 31.1 / 33.4 \quad 34.7 / 34.0 \quad 36.9 / 33.8 & 27.9 & 70.1 \\
YOLOE-26l-seg & 640 & Text / Visual & 36.8 / 33.7 & 37.8 / 36.3 & 35.1 / 37.6 \quad 37.6 / 36.2 \quad 38.5 / 36.1 & 32.3 & 88.3 \\
YOLOE-26x-seg & 640 & Text / Visual & 39.5 / 36.2 & 40.6 / 38.5 & 37.4 / 35.3 \quad 40.9 / 38.8 \quad 41.0 / 38.8 & 69.9 & 196.7 \\
\bottomrule
\end{tabular}
\end{table}

\begin{table}[h!]
\centering
\scriptsize
\setlength{\tabcolsep}{4pt}
\renewcommand{\arraystretch}{1.15}
\caption{YOLOE-26 \textbf{prompt-free} open-vocabulary instance segmentation performance at 640 px resolution. Results are reported on the minival set with end-to-end (e2e) evaluation. Prompt-free models use a built-in vocabulary (e.g., RAM++ tag set) and require no user-provided text or visual prompts. Models are trained on Objects365v1, GQA, and Flickr30k datasets. }
\label{tab:yoloe26_seg_pf}
\vspace{2pt}

\begin{tabular}{l c c c c c}
\toprule
\textbf{Model} & \textbf{Size} & \textbf{mAP$_{50\text{--}95}$} & \textbf{mAP$_{50}$} & \textbf{Params} & \textbf{FLOPs} \\
 & \textbf{(px)} & \textbf{(e2e)} & \textbf{(e2e)} & \textbf{(M)} & \textbf{(B)} \\
\midrule
YOLOE-26n-seg-pf & 640 & 16.6 & 22.7 & 6.5 & 15.8 \\
YOLOE-26s-seg-pf & 640 & 21.4 & 28.6 & 16.2 & 35.5 \\
YOLOE-26m-seg-pf & 640 & 25.7 & 33.6 & 36.2 & 122.1 \\
YOLOE-26l-seg-pf & 640 & 27.2 & 35.4 & 40.6 & 140.4 \\
YOLOE-26x-seg-pf & 640 & 29.9 & 38.7 & 86.3 & 314.4 \\
\bottomrule
\end{tabular}
\end{table}

Table~\ref{tab:yoloe26_seg_pf} evaluates the prompt-free setting, where no user-provided text or visual cues are available and the model relies on a built-in vocabulary for object discovery. As expected, absolute performance is lower than in the prompted setting, reflecting the increased difficulty of unconstrained open-world segmentation. Nevertheless, YOLOE-26x-seg-pf again demonstrates the strongest performance, achieving a mAP$_{50\text{-}95}$ of 29.9 and mAP$_{50}$ of 38.7, indicating robust object discovery capability without external prompts. This mode is particularly relevant for large-scale image or video understanding tasks such as autonomous exploration, content indexing, and scene parsing where manual prompt specification is impractical.

Importantly, the prompt-free results exhibit the same monotonic scaling behavior as the prompted setting, confirming that YOLOE-26’s architectural design generalizes across inference modes. Smaller prompt-free models, such as YOLOE-26n-seg-pf and YOLOE-26s-seg-pf, provide lightweight solutions for continuous background monitoring and edge-based perception, while larger variants enable richer semantic coverage in compute-rich environments. Overall, the combined analysis of Table~\ref{tab:yoloe26_seg_tv} and Table~\ref{tab:yoloe26_seg_pf} demonstrates that YOLOE-26 offers a flexible and scalable framework for real-world open-vocabulary instance segmentation, supporting both guided and unguided perception with strong accuracy–efficiency trade-offs suitable for diverse deployment scenarios.

\section{OPEN-VOCABULARY PROMPTING MECHANISMS}

A defining capability of YOLOE-26 is its support for \emph{open-vocabulary instance segmentation} through multiple prompting mechanisms \cite{xiao2026open, ma2025apovis, zhang2025low}. Unlike conventional YOLO-based detectors that rely on a fixed set of class logits learned during training, YOLOE-26 decouples semantic category reasoning from closed-set classification. Instead, object recognition is formulated as a similarity matching problem between learned object embeddings and prompt embeddings derived from text, visual examples, or a built-in vocabulary. This section describes the three complementary prompting mechanisms supported by YOLOE-26 text-prompted, visual-prompted, and prompt-free inference and explains how they collectively enable flexible, real-time, and deployment-friendly open-world perception.

\subsection{Text-Prompted Instance Segmentation}

Text prompting allows users to specify target objects using natural language descriptions such as \texttt{``person''}, \texttt{``bus''}, or \texttt{``red apple''}. In YOLOE-26, text prompts are encoded into semantic embeddings and aligned with object embeddings predicted at each anchor point. The resulting similarity scores determine both category assignment and instance mask generation.

To achieve efficient visual--textual alignment, YOLOE-26 adopts the \textbf{Re-parameterizable Region--Text Alignment (RepRTA)} strategy. During training, RepRTA introduces a lightweight auxiliary network that refines pretrained text embeddings to better align with object embeddings. Importantly, this auxiliary network is re-parameterized into the object embedding head after training, yielding \emph{zero additional inference cost}. This design preserves the speed and determinism of YOLOv26 while enabling open-vocabulary reasoning.

\begin{tcolorbox}[
  colframe=blue!70!black,
  colback=blue!5,
  title={Algorithm 1: Text-Prompted Instance Segmentation},
  fonttitle=\bfseries,
  arc=2mm
]
\small
\textbf{Input:} Image $I$, text prompts $\{t_1,\dots,t_C\}$ \\
\textbf{Output:} Instance masks and categories
\begin{enumerate}
  \item Extract multi-scale features using the YOLOv26 backbone and PAN.
  \item Predict object embeddings $\mathcal{O} \in \mathbb{R}^{N \times D}$.
  \item Encode text prompts as embeddings $\mathcal{P}_t \in \mathbb{R}^{C \times D}$.
  \item Compute similarity $\mathcal{O} \cdot \mathcal{P}_t^{\top}$.
  \item Assign categories and generate instance masks.
\end{enumerate}
\end{tcolorbox}

\noindent Example usage with text prompts:
\begin{lstlisting}
from ultralytics import YOLO

model = YOLO("yoloe-26l-seg.pt")
names = ["person", "bus"]
model.set_classes(names, model.get_text_pe(names))
results = model.predict("path/to/image.jpg")
results[0].show()
\end{lstlisting}

\subsection{Visual-Prompted Instance Segmentation}

Text descriptions are not always sufficient to specify objects precisely, especially in domains such as agriculture, medical imaging, or industrial inspection. YOLOE-26 therefore supports \textbf{visual prompting}, where users provide example bounding boxes or masks to define target objects.

This capability is enabled by the \textbf{Semantic-Activated Visual Prompt Encoder (SAVPE)}, which consists of two lightweight branches: (i) a semantic branch that extracts prompt-agnostic visual features, and (ii) an activation branch that encodes visual cues into prompt-aware weights. These branches are aggregated to form a compact visual prompt embedding aligned with object embeddings, enabling efficient matching with minimal computational overhead.

\begin{tcolorbox}[
  colframe=blue!70!black,
  colback=blue!5,
  title={Algorithm 2: Visual-Prompted Instance Segmentation},
  fonttitle=\bfseries,
  arc=2mm
]
\small
\textbf{Input:} Image $I$, visual prompts (boxes or masks) \\
\textbf{Output:} Instance masks of visually similar objects
\begin{enumerate}
  \item Extract multi-scale image features.
  \item Encode visual prompts using SAVPE.
  \item Generate visual prompt embedding $\mathcal{P}_v$.
  \item Match $\mathcal{P}_v$ with object embeddings $\mathcal{O}$.
  \item Predict instance masks for matched objects.
\end{enumerate}
\end{tcolorbox}

\noindent Example usage with visual prompts:
\begin{lstlisting}
import numpy as np
from ultralytics import YOLO
from ultralytics.models.yolo.yoloe import YOLOEVPSegPredictor

model = YOLO("yoloe-26l-seg.pt")

visual_prompts = dict(
    bboxes=np.array([[221.5, 405.8, 344.9, 857.5],
                     [120, 425, 160, 445]]),
    cls=np.array([0, 1])
)

results = model.predict(
    "path/to/image.jpg",
    visual_prompts=visual_prompts,
    predictor=YOLOEVPSegPredictor
)
results[0].show()
\end{lstlisting}

\subsection{Prompt-Free Instance Segmentation}

YOLOE-26 further supports a \textbf{prompt-free} inference mode for fully autonomous perception. These models operate using a built-in vocabulary of 4{,}585 categories derived from RAM++ tags. Rather than employing generative language models, YOLOE-26 introduces the \textbf{Lazy Region--Prompt Contrast (LRPC)} strategy, which first identifies object regions and then selectively retrieves category names only for relevant regions.

\begin{tcolorbox}[
  colframe=blue!70!black,
  colback=blue!5,
  title={Algorithm 3: Prompt-Free Instance Segmentation},
  fonttitle=\bfseries,
  arc=2mm
]
\small
\textbf{Input:} Image $I$ \\
\textbf{Output:} Instance masks and category names
\begin{enumerate}
  \item Predict objectness embeddings to locate candidate regions.
  \item Filter anchor points using a threshold.
  \item Lazily match filtered regions with the built-in vocabulary.
  \item Assign category labels and generate instance masks.
\end{enumerate}
\end{tcolorbox}

\noindent Example usage for prompt-free inference:
\begin{lstlisting}
from ultralytics import YOLO

model = YOLO("yoloe-26l-seg-pf.pt")
results = model.predict("path/to/image.jpg")
results[0].show()
\end{lstlisting}

Together, these three prompting mechanisms unify text-guided, example-guided, and autonomous perception within a single architecture. This versatility makes YOLOE-26 well suited for real-world deployment in robotics, autonomous systems, surveillance, industrial inspection, and precision agriculture, where object categories are dynamic and continuously evolving.

\section{TRAINING STRATEGY AND IMPLEMENTATION}
\label{sec:train_impl}

This section describes the training strategy and practical implementation details of YOLOE-26, with emphasis on how open-vocabulary learning is integrated into a YOLOv26-style real-time instance segmentation pipeline without sacrificing deployability. YOLOE-26 inherits the end-to-end, NMS-free efficiency of YOLOv26 while adopting the promptable open-vocabulary learning paradigm introduced in YOLOE \cite{liu2025yoloe}. The key idea is that the detector is trained not only to localize and segment objects, but also to produce \emph{semantic object embeddings} that can be matched against prompt embeddings derived from text, visual cues, or a built-in vocabulary. Unlike transformer-heavy open-vocabulary models that perform dense cross-attention between image tokens and text tokens at inference time, YOLOE-26 aims to push most of the cross-modal alignment complexity into training, then \emph{re-parameterize} modules so inference remains YOLO-like.

\subsection{Datasets, Annotation Sources, and Supervision Signals}
\label{subsec:datasets_supervision}

\textbf{Training data sources:} YOLOE-26 is trained using large-scale detection and grounding datasets that collectively provide diverse object categories, language grounding, and rich visual variability. In practice, three public sources are commonly used: Objects365 (object detection with bounding boxes), GQA (grounding-style annotations aligned with language), and Flickr30k Entities (phrase grounding). These datasets provide complementary supervision: Objects365 contributes broad object diversity and dense bounding boxes; GQA and Flickr30k connect text phrases to regions, which is crucial for learning prompt alignment.

\textbf{Segmentation supervision:} Since not all large-scale grounding/detection datasets provide high-quality instance masks, YOLOE-26 training typically uses \emph{pseudo-mask generation} to produce segmentation targets. A common approach is to generate instance masks from the provided bounding boxes using a strong segmentation model (e.g., SAM-family variants) and then refine the masks to reduce label noise. In implementation, the refinement step can include removing fragmented regions, suppressing out-of-box leakage, filtering small spurious components, and enforcing mask smoothness. The goal is not to replace human-annotated masks, but to create sufficiently accurate supervisory signals that enable training a segmentation head at scale.

\textbf{Multi-source supervision schema:} Let a training sample consist of an image $I$ with a set of annotated regions $\{(b_i, m_i, y_i)\}_{i=1}^{M}$, where $b_i$ is a bounding box, $m_i$ is an instance mask (ground-truth or pseudo-mask), and $y_i$ is a semantic label. In grounding datasets, $y_i$ may originate from a phrase rather than a canonical class name; YOLOE-26 therefore maps language phrases to prompt embeddings, allowing supervision to remain consistent even when vocabulary differs across datasets.

\textbf{Prompt-conditioned training targets:} For text prompting, each semantic label $y_i$ is represented by a text prompt embedding $p(y_i)\in\mathbb{R}^{D}$. For visual prompting, the training sample additionally includes a set of reference visual cues (e.g., a box or mask) used to construct a visual prompt embedding. For prompt-free training, the objective is to learn objectness and vocabulary retrieval without explicit user prompts. This is typically done in two stages: (i) learn a specialized \emph{objectness prompt} that identifies anchors corresponding to objects, and (ii) retrieve names from a built-in vocabulary by matching only those candidate anchors.

\subsection{Objective Functions and Optimization}
\label{subsec:loss_opt}

YOLOE-26 optimizes a multi-task loss that couples localization, segmentation, and semantic alignment. The model predicts, for each anchor point (or anchor-free location), a bounding box, a mask representation, and an object embedding. Let $\mathcal{O}\in\mathbb{R}^{N\times D}$ denote the predicted object embeddings for $N$ anchor points and embedding dimension $D$. Let prompt embeddings be $\mathcal{P}\in\mathbb{R}^{C\times D}$ for $C$ prompts. The similarity scores used for category assignment can be written as:
\begin{equation}
S = \mathcal{O}\,\mathcal{P}^{\top}\in\mathbb{R}^{N\times C},
\label{eq:sim_scores}
\end{equation}
where $S_{n,c}$ indicates how well anchor $n$ matches prompt $c$. These scores replace conventional fixed-class logits in closed-set detectors.

\textbf{(i) Embedding-based classification loss:} YOLOE-26 typically applies a binary cross-entropy style objective over similarity scores, treating correct anchor--prompt pairs as positives. Let $t_{n,c}\in\{0,1\}$ denote whether anchor $n$ matches prompt $c$. The classification term can be expressed as:
\begin{equation}
\mathcal{L}_{cls} = \frac{1}{NC}\sum_{n=1}^{N}\sum_{c=1}^{C} \mathrm{BCE}\big(\sigma(S_{n,c}),\, t_{n,c}\big),
\label{eq:cls_loss}
\end{equation}
where $\sigma(\cdot)$ is the sigmoid function. In practice, label assignment can follow a task-aligned matching strategy similar to modern YOLO training, ensuring stable gradients even under large prompt sets.

\textbf{(ii) Bounding box regression loss:} For localization, YOLOE-26 uses IoU-family losses (e.g., CIoU/GIoU) computed between predicted boxes $\hat{b}$ and ground-truth boxes $b$:
\begin{equation}
\mathcal{L}_{box} = \frac{1}{M}\sum_{i=1}^{M} \left(1-\mathrm{IoU}(\hat{b}_i, b_i)\right),
\label{eq:iou_loss}
\end{equation}
optionally with additional geometric penalties for aspect ratio and center distance. This term enforces tight spatial alignment and is essential for accurate instance masks.

\textbf{(iii) Localization refinement loss:} In segmentation-capable YOLO variants, distribution-based regression objectives are sometimes used for sub-bin precision. When included, the refinement term can be written as a distributional focal loss (DFL) style objective over discretized offsets. In YOLOv26-style deployments, regression objectives may be simplified for speed; therefore, YOLOE-26 implementations often expose this term as configurable depending on whether the goal is maximal accuracy or maximal efficiency.

\textbf{(iv) Mask segmentation loss:} YOLOE-26 employs a YOLACT/YOLO-Seg style mask representation that predicts a set of prototypes and per-instance mask coefficients. Let $M_i$ be the predicted mask for instance $i$ and $m_i$ the target mask. A standard choice is pixel-wise BCE:
\begin{equation}
\mathcal{L}_{mask} = \frac{1}{M}\sum_{i=1}^{M}\mathrm{BCE}(M_i, m_i),
\label{eq:mask_loss}
\end{equation}
optionally combined with dice loss for robustness to class imbalance in foreground/background pixels.

\textbf{(v) Total loss.} The overall objective is a weighted sum:
\begin{equation}
\mathcal{L} = \lambda_{cls}\mathcal{L}_{cls} + \lambda_{box}\mathcal{L}_{box} + \lambda_{mask}\mathcal{L}_{mask} + \lambda_{ref}\mathcal{L}_{ref},
\label{eq:total_loss}
\end{equation}
where $\lambda$ weights tune the balance between semantic alignment, localization, and segmentation quality.

\textbf{Optimization and schedules:} Training is commonly staged to reduce compute and stabilize optimization across prompt modes. A practical schedule is:
(1) \emph{Text-prompt pretraining}: learn strong region--text alignment with RepRTA and large-scale grounding data.
(2) \emph{Visual-prompt adaptation}: fine-tune SAVPE (often freezing most of the model) using box/mask-based visual cues, which reduces training cost because only a small prompt encoder is updated.
(3) \emph{Prompt-free specialization}: train an objectness prompt and enable lazy retrieval from a large vocabulary, emphasizing efficiency and coverage rather than prompt specificity.

Because RepRTA can be re-parameterized into the embedding head, the final inference graph can remain compact. For visual prompting, freezing most layers and updating only SAVPE is particularly beneficial: it reduces VRAM requirements, shortens training time, and preserves the pretrained detector representation. In practice, AdamW is often used for prompt encoder fine-tuning, while SGD-style optimizers may be used for large-scale pretraining depending on stability and throughput requirements.

\subsection{Implementation Pipeline, Trainers, and Ultralytics Integration}
\label{subsec:impl_ultralytics}

\textbf{Model variants and operating modes} YOLOE-26 is distributed in multiple scales (N/S/M/L/X) and in two operating families: (i) \emph{text/visual-prompt} models (e.g., \texttt{yoloe-26l-seg.pt}) and (ii) \emph{prompt-free} models (e.g., \texttt{yoloe-26l-seg-pf.pt}). Both families support inference, validation, training (fine-tuning), and export within the Ultralytics ecosystem, enabling consistent workflows from research to deployment.

\textbf{Fine-tuning on custom datasets} Fine-tuning YOLOE-26 on a custom segmentation dataset closely follows standard YOLO training, but requires a prompt-aware trainer to handle embedding-based classification and prompt construction. For instance segmentation fine-tuning, practitioners typically use a segmentation-specific trainer (e.g., \texttt{YOLOEPESegTrainer}) that ensures the mask branch, embedding head, and prompt modules are optimized consistently. Importantly, fine-tuning can be conducted either in closed-set style (fixed dataset categories) or in open-vocabulary style (dataset categories represented as prompts), depending on the desired deployment behavior.

\textbf{Detection-only fine-tuning from segmentation checkpoints} When training a detection model rather than segmentation, a practical approach is to initialize a detection configuration (YAML), load weights from a segmentation checkpoint of the same scale, and then train using a detection-specific trainer (e.g., \texttt{YOLOEPETrainer}). This reuses learned embeddings and localization features while discarding mask-specific parameters when unnecessary.

\textbf{Visual-prompt training efficiency} A distinctive implementation detail of YOLOE-style models is that visual-prompt models can be obtained by fine-tuning from trained text-prompt models. Since SAVPE is the main module requiring adaptation, one can freeze the entire backbone, neck, and most heads, and update only SAVPE-related layers. This significantly reduces compute, enabling short training runs that specialize the model for visual prompting. A typical engineering workflow is:
\begin{enumerate}
    \item Start from a well-trained text-prompt model checkpoint.
    \item (Optional) convert segmentation to detection graph for cheaper training.
    \item Freeze all layers except the SAVPE module.
    \item Train for a small number of epochs on visual-prompt supervision.
    \item Merge SAVPE back into the segmentation checkpoint if conversion was used.
\end{enumerate}
This workflow is particularly suitable for rapid iteration when new visual prompting behaviors are needed (e.g., one-shot part segmentation or instance retrieval in industrial scenes).

\textbf{Validation and prompt extraction} In validation, prompt embeddings must be constructed consistently with training. For example, visual prompt evaluation may require extracting visual embeddings for each dataset category from a reference set. Ultralytics-style APIs commonly support this with flags that automatically compute and cache category embeddings (e.g., loading visual prompt embeddings from a dataset). This design reduces user burden and standardizes evaluation protocols across datasets.

\textbf{Export and deployment} A major benefit of YOLOE-26 is that deployment follows familiar YOLO export pathways (ONNX, TensorRT, CoreML). For text-prompted export, practitioners configure the prompt set before exporting, so the exported model contains the folded prompt representation. This results in an inference graph compatible with edge runtimes, without requiring external text encoders at deployment. Prompt-free models export directly like standard YOLO models because they do not require runtime prompt inputs.

\textbf{Practical usage patterns} In real applications, prompt modes can be composed to maximize usability. A common pattern is \emph{prompt-free discovery} followed by \emph{prompted refinement}: the system first identifies a broad set of objects using the built-in vocabulary, then a user or downstream agent specifies a smaller set of target concepts through text or visual prompts to obtain precise instance masks. This hybrid workflow is effective in robotics, agricultural monitoring, and large-scale image/video analytics, where the object set evolves and the cost of repeated retraining is prohibitive.

Overall, YOLOE-26 training and implementation are designed around a deployment-first philosophy: large-scale open-vocabulary learning is achieved through prompt alignment objectives and staged specialization, while the final inference system remains lightweight, end-to-end, and compatible with standard YOLO acceleration toolchains.

\section{Conclusion and Future Roadmap}
\label{sec:conclusion_future}
This paper presented a systematic evaluation of YOLOE-26, a unified framework that integrates the deployment-oriented, NMS-free, end-to-end design of YOLOv26 with the open-vocabulary, promptable learning paradigm introduced by YOLOE as the foundational version. The resulting model advances open-vocabulary image segmentation by enabling real-time instance segmentation under three complementary modes: text-prompted, visual-prompted, and prompt-free inference. In contrast to transformer-heavy vision–language models that often incur substantial latency and memory overhead, YOLOE-26 preserves the speed and determinism of the YOLO family through embedding-based similarity matching and re-parameterizable prompting components. Overall, YOLOE-26 provides a practical balance between open-world semantic flexibility and edge-ready real-time segmentation, making it well aligned with high-impact application domains such as robotics, autonomous systems, precision agriculture, intelligent surveillance, medical imaging, and industrial inspection.

Despite these strengths, important limitations remain that motivate a clear roadmap for future research and engineering. First, prompt-free performance remains consistently below text- and visual-prompted settings, reflecting the intrinsic difficulty of unconstrained open-world object discovery in large-vocabulary environments. Second, large-scale training depends on multi-source supervision and pseudo-mask generation, which can introduce label noise and reduce boundary precision for thin structures, heavily occluded objects, and fine-grained categories. Third, open-vocabulary generalization is sensitive to prompt phrasing, dataset bias, and long-tail semantics, and the unified embedding space may not always fully separate visually similar categories without stronger alignment constraints. Fourth, while inference is efficient, practical deployment under extreme compute and energy constraints still poses challenges, including quantization robustness, memory overhead for large vocabularies, and reliable confidence calibration for safety-critical applications.


\begin{figure*}[ht!]
\centering
\resizebox{\linewidth}{!}{%
\begin{tikzpicture}[
    font=\Large,
    node distance=11mm and 16mm,
    every node/.style={draw, rounded corners, align=center, line width=0.7pt},
    root/.style={fill=gray!20, font=\bfseries\LARGE, minimum width=4.6cm, minimum height=1.30cm},
    pillar/.style={minimum width=4.4cm, minimum height=1.05cm, fill=white, font=\bfseries\Large},
    fut/.style={draw, rounded corners, line width=0.8pt, dashed, minimum width=4.2cm,
                minimum height=1.00cm, fill=white, font=\Large},
    edge/.style={->, line width=0.85pt},
    xlink/.style={dotted, line width=0.95pt},
    badge/.style={draw, rounded corners, line width=0.7pt, fill=white, font=\bfseries\Large,
                  minimum height=9mm, minimum width=4.6cm}
]

\node[root] (future) {YOLOE-26\\Future Roadmap};

\node[pillar, fill=blue!14, below left=of future, xshift=-12mm] (eff)
{Edge-Efficient\\Open-Vocab Seg};
\node[pillar, fill=orange!16, below=of future] (rel)
{Reliable \&\\Robust Prompting};
\node[pillar, fill=green!16, below right=of future, xshift=12mm] (agi)
{Agentic AI\\\& Autonomous Perception};

\draw[edge] (future) -- (eff);
\draw[edge] (future) -- (rel);
\draw[edge] (future) -- (agi);

\node[fut, below=of eff] (e1) {Quantization-aware\\training (INT8/FP8)};
\node[fut, below=of e1] (e2) {Prompt caching \&\\kernel folding (RepRTA)};
\node[fut, below=of e2] (e3) {Vocabulary indexing\\(ANN retrieval for prompts)};
\node[fut, below=of e3] (e4) {Streaming video:\\temporal mask propagation};
\node[fut, below=of e4] (e5) {Distillation to\\tiny YOLOE-26 (N/S)};
\node[fut, below=of e5] (e6) {Hardware-aware export:\\ONNX/TensorRT/CoreML};

\draw[edge] (eff) -- (e1);
\draw[edge] (e1) -- (e2);
\draw[edge] (e2) -- (e3);
\draw[edge] (e3) -- (e4);
\draw[edge] (e4) -- (e5);
\draw[edge] (e5) -- (e6);

\node[fut, below=of rel] (r1) {Prompt robustness:\\synonyms \& paraphrases};
\node[fut, below=of r1] (r2) {Uncertainty-aware\\thresholding \& abstention};
\node[fut, below=of r2] (r3) {Long-tail calibration:\\rare/common/frequent};
\node[fut, below=of r3] (r4) {Domain shift handling:\\weather, blur, low-light};
\node[fut, below=of r4] (r5) {Noisy pseudo-mask\\robust training (SAM-based)};
\node[fut, below=of r5] (r6) {Safety checks:\\false-positive control \& QA};

\draw[edge] (rel) -- (r1);
\draw[edge] (r1) -- (r2);
\draw[edge] (r2) -- (r3);
\draw[edge] (r3) -- (r4);
\draw[edge] (r4) -- (r5);
\draw[edge] (r5) -- (r6);

\node[fut, below=of agi] (a1) {Perception agent:\\discover $\rightarrow$ refine prompts};
\node[fut, below=of a1] (a2) {Active vision:\\reframe, zoom, multi-view consistency};
\node[fut, below=of a2] (a3) {Memory for prompts:\\episodic + vector cache};
\node[fut, below=of a3] (a4) {Self-training loop:\\selective pseudo-labeling};
\node[fut, below=of a4] (a5) {Tool use \& pipelines:\\auto-annotate \& audit};
\node[fut, below=of a5] (a6) {Continual learning:\\new classes without forgetting};

\draw[edge] (agi) -- (a1);
\draw[edge] (a1) -- (a2);
\draw[edge] (a2) -- (a3);
\draw[edge] (a3) -- (a4);
\draw[edge] (a4) -- (a5);
\draw[edge] (a5) -- (a6);

\draw[xlink] (e3.east) -- ++(10mm,0) |- (r3.west);
\draw[xlink] (e4.east) -- ++(10mm,0) |- (r4.west);

\draw[xlink] (r2.east) -- ++(10mm,0) |- (a1.west);
\draw[xlink] (r5.east) -- ++(10mm,0) |- (a4.west);

\draw[xlink] (e2.south) -- ++(0,-8mm) -| (a3.south);
\draw[xlink] (e6.south) -- ++(0,-8mm) -| (a6.south);

\node[badge, below=18mm of r6] (goal)
{Goal: Fully Real-Time Open-Vocabulary Instance Segmentation on Edge};

\draw[xlink] (e6.south) -- ++(0,-10mm) -| (goal.west);
\draw[xlink] (r6.south) -- (goal.north);
\draw[xlink] (a6.south) -- ++(0,-10mm) -| (goal.east);

\end{tikzpicture}
}

\caption{\textbf{Future roadmap for YOLOE-26 and beyond.}
Dashed boxes summarize prioritized research directions across three coupled thrusts:
(i) \emph{Edge-Efficient Open-Vocabulary Segmentation} (deployment, compression, fast retrieval),
(ii) \emph{Reliable \& Robust Prompting} (calibration, domain shift, noisy supervision),
and (iii) \emph{Agentic AI \& Autonomous Perception} (prompt refinement, memory, continual learning).
Dotted links highlight dependencies among efficiency, reliability, and agentic autonomy.}
\label{fig:yoloe26_future}
\end{figure*}
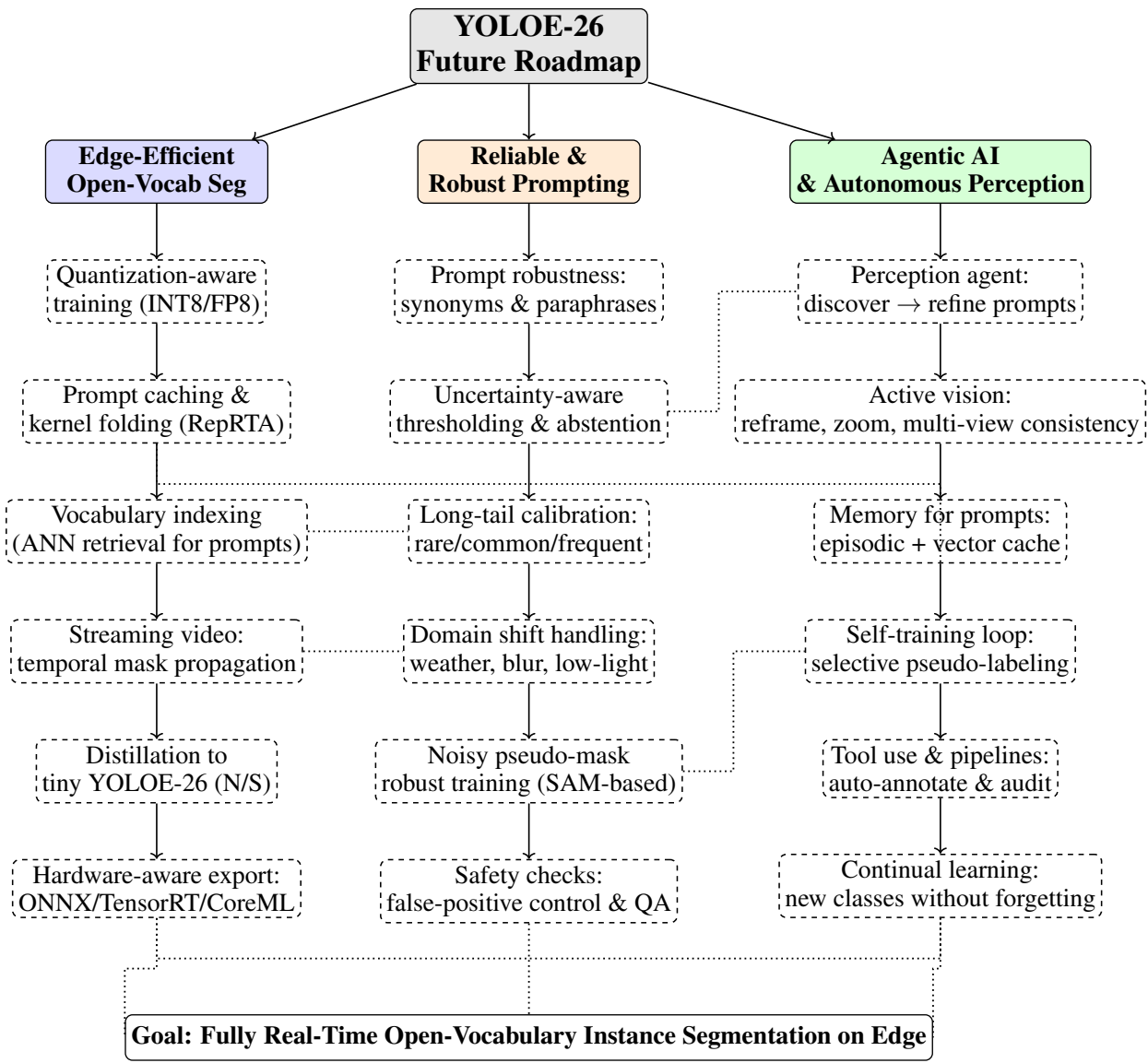

Future advancements of YOLOE-26, as illustrated in Figure~\ref{fig:yoloe26_future}, should prioritize fully autonomous open-vocabulary segmentation through agentic AI and AI agents that continuously improve perception in real-world environments. A natural direction is to integrate YOLOE-26 into an agentic perception loop that performs prompt-free object discovery, automatic prompt refinement using contextual cues, uncertainty-aware re-prompting when predictions are ambiguous, and self-verification through temporal or multi-view consistency checks. Such an agentic learning pipeline can enable closed-loop semantic adaptation without repeated full retraining, supporting long-term autonomy in dynamic and evolving scenes. In parallel, incorporating lightweight vision–language reasoning modules can help generate more discriminative prompts, such as attribute-based or compositional descriptions, improving robustness to domain shift and long-tail categories.

From a learning and systems perspective, several roadmap directions can further strengthen YOLOE-26 as a general-purpose open-vocabulary segmentation framework. Continual and federated open-vocabulary learning can enable incremental category expansion and personalized adaptation while mitigating catastrophic forgetting and preserving data privacy. Self-supervised and weakly supervised mask refinement strategies, including iterative pseudo-label cleanup, boundary-aware losses, and temporal consistency in video, can reduce reliance on costly manual annotations and improve pixel-level accuracy. Hierarchical and compositional embedding spaces can explicitly encode categories, attributes, and object parts, improving separability for visually similar classes and supporting scalable prompt-free retrieval. Finally, edge-first deployment optimization, including quantization-aware training, model distillation, prompt caching, and efficient indexing for large vocabularies, will be critical to ensure real-time performance on CPUs, embedded GPUs, and mobile NPUs.

In summary, YOLOE-26 demonstrates that open-vocabulary image segmentation can be achieved with YOLO-level efficiency by tightly integrating end-to-end detection with lightweight vision–language embedding mechanisms. Looking ahead, the convergence of YOLOE-26 with AI agents and agentic learning paradigms offers a compelling pathway toward fully autonomous, self-improving, and deployment-ready vision–language segmentation systems capable of discovering new objects, refining semantic understanding, adapting to evolving environments, and maintaining real-time performance across diverse real-world scenarios.

\bibliographystyle{unsrt}  
\bibliography{references}

\end{document}